\def\year{2021}\relax
\documentclass[letterpaper]{article} 
\usepackage{aaai21}  
\usepackage{times}  
\usepackage{helvet} 
\usepackage{courier}  
\usepackage[hyphens]{url}  
\usepackage{graphicx} 
\usepackage{natbib}  
\usepackage{caption} 
\usepackage{booktabs} 
\usepackage{multirow} 
\usepackage{color} 
\usepackage{algorithm}
\usepackage{algorithmic}
\usepackage{amsmath}
\usepackage{booktabs}  
\usepackage{threeparttable}  
\usepackage{tabularx}
\usepackage{adjustbox}
\usepackage{wasysym}

\usepackage{comment}

\title{ Unstructured Knowledge Access in Task-oriented Dialog Modeling using Language Inference, Knowledge Retrieval and Knowledge-Integrative \\Response Generation}

\author{
Mudit Chaudhary$^{1}$\footnote{All authors have contributed equally.}, Borislav Dzodzo$^{1}$, Sida Huang$^{1}$, Chun Hei Lo$^{1}$, Mingzhi Lyu$^{1}$, \\Lun Yiu Nie$^{1}$, Jinbo Xing$^{1}$, Tianhua Zhang$^{2}$, Xiaoying Zhang$^{1}$, Jingyan Zhou$^{1}$,\\ Hong Cheng$^{1,2}$, Wai Lam$^{1,2}$, Helen Meng$^{1,2}$ \\\vspace{0.1cm}
}
\affiliations{
    $^{1}$The Chinese University of Hong Kong\\
    $^{2}$Centre for Perceptual and Interactive Intelligence\\
    \vspace{0.1cm}
    muditchaudhary@cuhk.edu.hk, \\
    \{bdzodzo, chlo, zhangxy, jyzhou, hcheng, wlam, hmmeng\}@se.cuhk.edu.hk \\
    \{sdhuang, lynie, thzhang\}@link.cuhk.edu.hk, 
    \{mzlyu, jbxing\}@cse.cuhk.edu.hk, \\
}
\begin{document}
 
\maketitle

\begin{abstract}

Dialog systems enriched with external knowledge can handle user queries that are outside the scope of the supporting databases/APIs. In this paper, we follow the baseline provided in DSTC9 Track 1 and propose three subsystems, KDEAK, KnowleDgEFactor, and Ens-GPT, which form the pipeline for a task-oriented dialog system capable of accessing unstructured knowledge. Specifically, KDEAK performs knowledge-seeking turn detection by formulating the problem as natural language inference using knowledge from dialogs, databases and FAQs. KnowleDgEFactor accomplishes the knowledge selection task by formulating a factorized knowledge/document retrieval problem with three modules performing domain, entity and knowledge level analyses.
Ens-GPT generates a response by first processing multiple knowledge snippets, followed by an ensemble algorithm that decides if the response should be solely derived from a GPT2-XL model, or regenerated in combination with the top-ranking knowledge snippet. Experimental results demonstrate that the proposed pipeline system outperforms the baseline and generates high-quality responses, achieving at least 58.77\% improvement on BLEU-4 score. 
\end{abstract}

\section{Introduction}
By incorporating the external knowledge sources available on webpages, task-oriented dialog systems can be empowered to handle various user requests that are outside the coverage of their APIs or databases. Therefore, we set out to create a dialog system that outperforms the Ninth Dialog System Technology Challenge (DSTC9) Track 1 baseline \cite{kim2020domain,gunasekara2020overview}. The baseline method is a pipeline composed of three tasks: the first task recognizes if a dialog response requires knowledge outside of a provided MultiWOZ 2.1 database \cite{eric2019multiwoz}. If so, the second task then retrieves the most relevant knowledge snippets from an external knowledge base, which are subsequently used together with the dialog context to generate a response in the third task. Specifically, all the three tasks are handled by the variants of pre-trained GPT2 models \cite{vaswani2017attention, Wolf2019HuggingFacesTS}. 

Formally, the external knowledge base $K$ is composed of knowledge snippets $k_1,\ldots, k_n$. $D$ is the set of all domains. For the DSTC9 Track 1 Training Set, $D=\{\text{hotel},\text{restaurant},\text{train},\text{taxi}\}$. Table \ref{tab:knowexample} shows examples of the two types of knowledge, namely a domain-wide knowledge snippet directly under a specific domain $d_i=\text{train}$, and an entity-specific knowledge snippet of entity $e_i=\text{Avalon}$, which belongs to the domain hotel. $D_w$ and $D_e$  refer to the domains that contain only domain-wide and only entity-specific knowledge snippets respectively, $D_w \cup D_e = D$ and $D_w \cap D_e = \emptyset$. A snippet $k_i$ consists of a title (question) and a body (answer). A knowledge snippet $k_i$ is considered \textit{in-domain (ID)} if its domain $d_i$ was seen during the training of the models; otherwise, it is considered \textit{out-of-domain (OOD)}. 
The dialog history $U_t=\{u_{t-w+1},\ldots, u_{t-1},u_t\}$ contains utterances $u_i$ where $t$ is the time step of the current user utterance and $w$ is the size of dialog context window. Responses to this dialog are found in the ground truth ${r}_{t+1}$ or they can be generated by our system $\Tilde{r}_{t+1}$.

\begin{table}[!t]
  \centering
  \begin{threeparttable}
  \fontsize{9}{10}
  \selectfont
    \begin{tabular}{lp{0.11\columnwidth}p{0.6\columnwidth}}
        \toprule
         \textbf{Domain} & \textbf{Entity} & \textbf{Snippet}\cr
        \midrule
        \multirow{2}*{Train}& \multirow{2}*{--}
        &\textbf{T}: Is there a charge for using WiFi?\cr
        &&\textbf{B}: Wifi is available free of charge.\cr
        \midrule
        \multirow{2}*{Hotel}& \multirow{2}{0.11\columnwidth}{Avalon}
        &\textbf{T}: Are pets allowed on site?\cr
        &&\textbf{B}: Pets are not allowed at avalon.\cr
        \bottomrule
    \end{tabular}
  \end{threeparttable}
  \caption{Example of a domain-wide (line-1) and an entity-specific knowledge snippet (line-2). $T, B$ represent the title and the body.}
  \label{tab:knowexample}
\end{table}
We created a transparent, factorized, generalisable and knowledge-grounded task-oriented conversational system with code available at \textit{http://bit.ly/2ISy3KW}.   Multiple information retrieval hypotheses are considered when constructing the response and this significantly improves results.
When the three tasks are integrated they significantly outperform the baseline in terms of automated metrics.

\begin{figure}
\centering
\includegraphics[width=0.75\columnwidth]{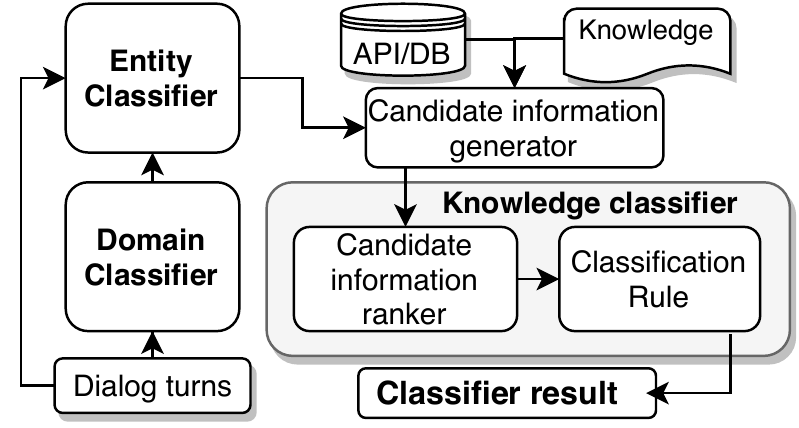}
\caption{KDEAK pipeline}
\label{KDEAK_pipeline_diagram}
\end{figure}

\begin{table*}[!t]
 \fontsize{9}{9}
 \selectfont
 \centering
 \begin{threeparttable}
    \begin{tabular}{lp{0.3\linewidth}p{0.25\linewidth}p{0.18\linewidth}}
    \toprule  
    \textbf{Speaker} & \textbf{Utterance}  ($u_t$) & \textbf{Knowledge Snippet} ($k_i$) & \textbf{Database Entry}\cr
    \midrule
    Assistant&Would you like to book the SW hotel? &- &\textit{name:SW Hotel}\\
    &&&\textit{address:615 Broadway}\cr
    User&Yes, I can reach SW hotel by taxi. What breakfast options are available there?&\textbf{T:} Does SW Hotel offer breakfast? \textbf{B:} No, we don't offer breakfast.&\textit{postcode:94133 
    type:Hotel}\\
    \bottomrule  
    \end{tabular}
 \end{threeparttable}
 \caption{Excerpt of last 2 dialog turns from \textit{hotel} domain with relevant knowledge snippet (T: title, B: body) and database entry. } 
 \label{tab:Task1_example}
\end{table*}

\section{Methodology}
\subsection{Task 1 -- Knowledge-seeking Turn Detection}
As mentioned earlier, Task 1 classifies whether information from the database or external knowledge is required to answer a user's query.

We introduce KDEAK (\textbf{K}nowledge-seeking turn detection using \textbf{D}omain, \textbf{E}ntity, \textbf{A}PI/DB and \textbf{K}nowledge) shown in Figure \ref{KDEAK_pipeline_diagram}. The domain classifier helps the entity classifier determine the dialog's relevant entity. We generate candidate information snippets from the selected entity's database and knowledge. The knowledge classifier ranks and classifies the candidate information snippets to determine whether the database or knowledge answers the user's query. In the subsequent sections, we illustrate our KDEAK's modules using the example from Table \ref{tab:Task1_example}. What differentiates KDEAK from Task 2 is that its Domain Classifier can identify domains in the non-knowledge-seeking turns and the Knowledge Classifier's ability to select the relevant API/DB information.

\begin{table*}[!t]
    \fontsize{10}{10}
    \selectfont
  \centering
  \begin{threeparttable}
    \begin{tabular}{lcccccccc}
    \toprule  
    \multirow{2}{*}{\textbf{Method}}& 
    \multicolumn{4}{c}{\textbf{Validation Set}}&\multicolumn{4}{c}{\textbf{Test Set}}\cr  
     \cmidrule(lr){2-5} \cmidrule(lr){6-9}
    &Accuracy&Precision&Recall&F1&Accuracy&Precision&Recall&F1\cr 
    \midrule
    GPT2-Baseline&\textbf{0.995}&\textbf{0.999}&0.982&\textbf{0.991}&0.946&\textbf{0.993}&0.892&0.940\cr
    KDEAK\textsuperscript{$\wedge$}&0.993&0.980&\textbf{0.993}&0.986&0.924&0.989&0.849&0.914\cr
    KDEAK\textsuperscript{*}&0.994&0.993&0.986&0.989&\textbf{0.971}&0.985&\textbf{0.952}&\textbf{0.968}\cr
    \bottomrule  
    \end{tabular}
  \end{threeparttable}
  \caption{Evaluation results of Task 1 on knowledge-seeking turn detection on DSTC9 Track 1 Validation and Test Sets. \textsuperscript{$\wedge$} Submitted system using $N_{dialog}=1$ without premise template. \textsuperscript{*} Improved system using $N_{dialog}=2$ and premise template.} 
  \label{tab:KDEAK-performance}
\end{table*}

\paragraph{NLI Problem Formulation.}
We formulate Task 1's domain classification, and knowledge classification problems as a Natural Language Inference (NLI) problem \cite{10.1007/11736790_9}. The NLI problem deals with a pair of statements -- hypothesis and premise. Given the premise, it determines whether the hypothesis is True (i.e., an entailment), False (i.e., a contradiction), or Undetermined. For example, if ``I want to book a hotel" is the premise, the hypotheses ``The user wants to book a hotel"  is True and ``The user wants to book a taxi" is False. We leverage a pre-trained NLI model \cite{lewis2019bart} for classification in Task 1. We use the last $N_{dialog}$ turns for premise generation. We pair each premise with a set of generated candidate hypotheses using domain and knowledge labels. We find the NLI approach more robust against unseen domains as compared to the baseline.

\paragraph{Module 1 - Domain Classifier.}
This module classifies the dialog turn's relevant domain. We generate the premise using the following premise template -- ``Assistant says \$system\_response. User says \$user\_response" in each dialog turn to distinguish between user and system response ($N_{dialog}=2$). Based on the example in Table \ref{tab:Task1_example}, the premise will be \textit{``Assistant says The SW ... book? User says Yes ... there?"} We pair the premise with a candidate hypothesis for each domain $d_i \in D$ using the hypothesis template -- \textit{``The user is asking about $d_i$."}. We feed these pairs into the NLI model to find the most probable domain by performing a softmax on each candidate hypothesis' output entailment probability. The domain $d_i$ with the maximum entailment probability is selected for the dialog turn. 

We use Bidirectional and Auto-Regressive Transformers (BART) model \cite{lewis2019bart} initially fine-tuned on MultiNLI \cite{williams-etal-2018-broad}. We further fine-tune our model on MultiWOZ2.2 \cite{zang-etal-2020-multiwoz} and DSTC9 Track 1 Training Set on all eight domains of MultiWOZ2.2. For training, we generate the premise and hypotheses using the templates mentioned above. Each premise with ground-truth domain $d_i$ is paired with the hypothesis corresponding to $d_i$ and marked as entailment. We also pair the same premise with the remaining $|D| - 1$ hypotheses and mark them as contradiction.  For inference, we use Huggingface's \cite{Wolf2019HuggingFacesTS} `classification-as-NLI' based zero-shot-classification pipeline.

\paragraph{Module 2 - Entity Classifier.}
The Entity Classifier uses the selected domain from the Domain Classifier to further process the dialog turn in focus. We devise a Surface Matching Algorithm (SMA) to match the possible entities within the dialog history with carefully designed heuristics, based on the intuition that the later the entity appears in the dialog history, the more likely it is the target. Approximate string matching is also incorporated into the algorithm to enhance its robustness to alias matching and misspelling. For instance, SMA is capable of retrieving the entities \textit{A \& B Guest House} from seeing `A and B', \textit{Avalon} from seeing `Avolon.' The selected domain label helps reduce the entity search space.  Following on with our example (see Table 2), the entities corresponding to the \textit{hotel} domain are searched to see if they occur in the dialog turn.  Consequently, the matching algorithm identifies \textit{SW Hotel} as the entity.

\paragraph{Module 3 - Candidate Information Generator (CIG).}
Given the identified entity from the Entity Classifier, this module consolidates the relevant database snippets and knowledge snippets for the entity and places them into an information candidate pool $C_{entity}$ which will be used by the Knowledge Classifier in the subsequent step. As we observe in Table \ref{tab:Task1_example}, database snippets are not natural sounding like knowledge-snippets, so we pre-process them using suitable formatting templates before adding them to $C_{entity}$. Based on Table \ref{tab:Task1_example}'s example, the database snippet -- \textit{\{name: SW Hotel, postcode: 94133\}} becomes \textit{``Postcode for SW Hotel is 94133."} We also add pseudo-candidates to $C_{entity}$ to deal with cases where information is not present in either database or knowledge, e.g., ``Goodbye", ``I want to book a hotel", ``Thanks", etc. Following on with our example, CIG consolidates SW Hotel's database and knowledge snippets and pseudo-candidates into the candidate pool $C_{SW\_hotel}$. 

\paragraph{Module 4 - Knowledge Classifier.}
This module consists of an NLI-based ranker for the candidates in the candidate information pool $C_{entity}$ output by the Candidate Information Generator. 
The dialog turn in focus is used as the premise and each information candidate in the pool $c_i \in C_{entity}$ is used as the hypothesis to form $|C_{entity}|$ premise-hypothesis pairs.  The candidates are ranked according to the entailment probabilities. The final classification rule classifies the turn as knowledge-seeking or not:
\begin{equation}
  label=\begin{cases} False& \text{if } c^{top}_i \notin K\\ True &otherwise \end{cases}
\end{equation}
where $c^{top}_i$ is the top-ranking candidate and $K$ is the set of knowledge-snippets. Following on with the example (See Table \ref{tab:Task1_example}), the latest user turn ``Yes, I ... available there?" is used as the premise ($N_{dialog}=1$) and it is paired with each candidate $c_i \in C_{SW\_hotel}$. The pairs are fed into the fine-tuned NLI-model to rank the snippets. The top-ranking candidate -- \textit{``No, we don't offer breakfast."} is a knowledge snippet. Hence, the example user's dialog turn is finally classified as knowledge-seeking.

As regards implementation, we use the previously mentioned BART model as the base model for the \textit{Candidate Information Ranker}. As the DSTC9 Track 1 Training Set provides database labels for non-knowledge-seeking turns, we generate pseudo-database-labels for the non-knowledge-seeking turns using an NLI BART model fine-tuned on MultiWoz 2.1. We fine-tune our ranker model on the DSTC9 Track 1 Training Set using these pseudo-labels and knowledge-seeking examples. We follow the same training example sampling, and inference processes as adopted for the Domain Classifier, but without the hypothesis and premise templates.

\subsection{Task 2 -- Knowledge Selection}
Once a user turn is determined to be knowledge-seeking by Task1, Task 2 selects the relevant knowledge snippets $k_i$ from the external knowledge base $K = \{k_1, \ldots, k_n\}$ based on a dialog history $U_t$. While one or more knowledge snippet(s) may be fitting for an answer, only one is considered most relevant and correct in the DSTC9 Track 1 challenge \cite{kim2020domain}.

\subsubsection{A Factorized Approach}
We model Task 2 as a knowledge retrieval (or more specifically, document retrieval) problem, i.e., given the query dialog history $U_t$, we retrieve the most relevant knowledge snippet $k_i \in K$ from the set of all knowledge snippets $K$ ranked by a function $f$. In this context, the function $f$ is the probability of selecting a knowledge snippet $k_i \in_d d_i\text{ where } d_i \in D_w$, or $k_i \in_e e_i \text{ where } e_i \in_d d_i, d_i \in D_e$ ($\in_e$ and $\in_d$ denote the relations `belongs to the domain' and `belongs to the entity' respectively). Therefore, the selected knowledge snippet $k_i \in_d d_i$ is given by:
\begin{align}
\mathop{\arg\max}_{k_i} f(k_i\mid U_t)  = \mathop{\arg\max}_{k_i \in_d d_i} P(d_i, k_i\mid U_t)
\end{align}

We propose that we first recognize the possible target domains $d_i \in D_w$ and entities $e_i \in_d d_i \text{ where } d_i \in D_e$ and estimate the relevance of the domains to the dialog history before choosing the appropriate knowledge snippet, since it can drastically narrow the search space for knowledge snippets. In other words, factorization reduces the problem of Task 2 into three sub-tasks, for each of which models can be trained for target discrimination. 
Consequently, we have:
\begin{align}
\mathop{\arg\max}_{k_i \in_d \{d_i: d_i \in D'\}} P(d_i \mid U_t) P(k_i \mid d_i, U_t) 
\end{align}
where $D' = \{d_i: d_i \in D_w, d_i \in O_{DE}\} \cup \{d_i: e_i \in_d d_i, d_i \in D_e, e_i \in O_{DE}\}$ ($O_{DE}$ refers to the output of Module 1, which is the set of extracted domains and entity candidates). $P(d_i \mid U_t)$ and $P(k_i \mid d_i, U_t)$ are estimated using Modules 2 and 3 respectively. The three modules are described in the following sections.

\subsubsection{BERT Backbone}
The computation of the factored probabilities $P(k_i \mid d_i, U_t)$ and $P(d_i \mid U_t)$ naturally resorts to natural language understanding (NLU) models. We employ BERT \cite{devlin2018bert} as the NLU backbone and propose KnowleDgEFactor (A \textbf{Factor}ized Approach to \textbf{D}omain, \textbf{E}ntity and \textbf{K}nowledge Selection). Three neural models are developed -- the BERT for Domain \& Entity Model (BERT-DE) in Module 1, BERT for Domain Model (BERT-D) in Module 2 and BERT for Knowledge Model (BERT-K) in Module 3.

\subsubsection{Module 1 - Domain and Entity Selection. }
We use the heuristics-based surface matching algorithm SMA (described in Section 2.1) to match the possible domains $d_i \in D_w$ and entities $e_i \in_d d_i \text{ where } d_i \in D_e$.

In view of the high generalization power of neural models, we propose a domain-entity classifier (BERT-DE) to refine the results obtained by SMA.
 
A dialog history $U_t$ is concatenated with a domain $d_i$ (and an entity $e_i \in_d d_i$ if $d_i \in D_e$) as the input to the BERT-DE. 
For example, \textit{train} ($\in D_w$ ) and \textit{hotel} ($\in D_e$ ) concatenated with \textit{Autumn House} ($\in_e$ `hotel') are the two kinds of input. BERT-DE computes the probability that the dialog history $U_t$ is relevant to each domain ($\in D_w$ ) or entity ($\in_e d_i$ where $d_i \in D_e$) and outputs the top-1 result with the highest probability, which is then added to the candidates if it has a different domain than that of the top-1 retrieved by SMA. In the end, we keep at most one entity per domain and finally output $O_{DE}$ for Module 3.

\subsubsection{Module 2 - Domain Probability Estimation. }

BERT-D is a multi-class domain classifier. Given the concatenation of a dialog history $U_t$ and a domain $d_i$ (e.g., \textit{hotel, train, etc.}) as input, it estimates and outputs $P(d_i \mid U_t)$, the probability that $U_t$ is relevant to $d_i$.

We combine the DSTC9 Track 1 Training Set with the MultiWOZ2.2 Data Set \cite{zang-etal-2020-multiwoz} to fine-tune the BERT-D on eight domains, i.e., \textit{hotel}, \textit{restaurant}, \textit{train}, \textit{taxi}, \textit{attraction}, \textit{hospital}, \textit{police} and \textit{bus}, to make the model more generalized and robust. 

The BERT-D differs from the domain classifier in Task 1 since we only focus on knowledge-seeking turns whereas Task 1's model needs to be applied to both knowledge-seeking and non-knowledge-seeking turns. Examples for the two cases are as follow:

\begin{itemize}
  \item [] 
    Case 1 - Non-knowledge-seeking Turn
    \begin{itemize}
        \item[]
         User: I am looking for an expensive indian restaurant in the area of centre.
         \item[] 
         Task 1 Domain Classifier: restaurant.
         \item[]
         Task 2 Domain Classifier: N/A (ignore the turn).
    \end{itemize}
    \item [] 
    Case 2 - Knowledge-seeking Turn
    \begin{itemize}
        \item[]
         User: Does this hotel offer its guests wifi services?
         \item[] 
         Task 1 Domain Classifier: hotel.
         \item[]
         Task 2 Domain Classifier: hotel.
    \end{itemize}

\end{itemize}

\subsubsection{Module 3 - Knowledge Probability Estimation. }
The BERT-K is designed to estimate $P(k_i \mid d_i,U_t)$ for all knowledge snippets $k_i$ of the domains and entities selected in Module 1. As most of the users' queries are embedded in the current user turn, the input to the BERT-K is the concatenation of the current user utterance $u_t$, a domain $d_i$ and a knowledge snippet $k_i$ (title \& body). For both the current user utterance and the knowledge snippet, any matched entity name is replaced by its domain so that the model only focuses on the semantics of the query but not any information about the entity, which has already been processed by previous modules.


\subsection{Task 3 -- Knowledge-grounded Response Generation}
Task 3 takes a knowledge-integrative approach to generate a system response based on the dialog history $U_t$ and the top-$k$ ranking knowledge snippets $k_i$ based on their confidence values $p_i$, which are provided by Task 2.
We develop an ensemble system Ens-GPT that incorporates two different approaches to deal with the two scenarios (ID and OOD). If the domain of the top knowledge snippet was seen in training then response generation will be conducted as ID and otherwise as OOD.  For ID cases with available training data, we adopt a \textit{Neural Response Generation} approach.  
For OOD cases, we adopt a retrieval-based approach referred to as \textit{Neural-Enhanced Response Reconstruction}. 

\subsubsection{Neural Response Generation.}

Our neural response generation approach \textit{GPT2-XL with multi-knowledge snippets} (GPT2-XL for short) follows the DSTC9 Track 1 baseline neural generation model in \cite{kim2020domain} to leverage the large pre-trained language model GPT2. 
The baseline neural generation model uses the ground truth knowledge snippet and dialog history $U_t$ as input for fine-tuning GPT2 \textit{small}, and the ground truth response $r_{t+1}$ as target. During testing, the baseline model uses knowledge from the top-ranking snippet output by Task 2.

As GPT2 \textit{XL} has a greater number of parameters to capture more information, we adopt the much larger pre-trained model GPT2 \textit{XL} other than the GPT2 \textit{small} used in the baseline model.
We should note that the actual correct knowledge snippet may not always rank top in the shortlisted snippets from Task 2, but most of the time they lie within the top 5 retrieved snippets. Hence, we use multiple knowledge snippets in the input, $n$ in total. For model fine-tuning, besides the ground truth snippet, we also randomly select $n-1$ additional snippets that have the same domain and entity with the ground truth snippet and append them to the input. Correspondingly, we use top-$n$ snippets from the retrieved top-ranking snippet list from Task 2 in the input for evaluation.

\subsubsection{Neural-Enhanced Response Reconstruction.}
Typical responses may consist of two parts -- (i) an informative \textit{body} which answers to the user's query; and (ii) a \textit{prompt} to move the dialog forward.  For example:
\begin{itemize}
  \item [] 
User: “Does this hotel allow children to stay there?”
  \item [] 
  Ground Truth Response: “Kids \textbf{of all ages} are welcome as guests of this establishment. \textit{Do you want to proceed with the booking?}”
\end{itemize}

Since the knowledge snippets made available are derived from FAQs, the top snippet is used as the \textit{body} in the response. 
Therefore, the GPT2-XL Response Reconstruction (GPT2-XL-RR) method forms an informative and accurate response by replacing the body of the neural generated response with the top-ranking snippet, while preserving the prompt in the generated response. For example, given:

\begin{itemize}
  \item [] 
Top-ranking knowledge snippet: “Children \textbf{of any age} are welcome at The Lensﬁeld Hotel.”
    \item []
GPT2-XL generated response: “Yes, The Lensﬁeld Hotel welcomes children to stay. \textit{Should I make the reservation now?}”
  \end{itemize}
The GPT2-XL-RR constructs the response as “Children \textbf{of any age} are welcome at The Lensﬁeld Hotel. \textit{Should I make the reservation now?}”

\subsubsection{Ensemble System.}
To utilize the two approaches above, a decision tree is designed for the ensemble system Ens-GPT. 
The system first checks if the user query is ID or OOD, which is detected by Task 2 and indicated by the domain of the top-ranking retrieved snippet.
For ID cases with available training data, the neural model GPT2-XL is generally well-trained, so it can generate relevant responses to the dialog even when the correct retrieved snippet is not retrieved. Therefore, given ID user queries, GPT2-XL is used for response generation.

On the other hand, if the current user query is OOD, the ensemble's heuristic will check if the top-ranking knowledge snippet has a sufficiently high confidence value $p$ (which is empirically set as 5x of the confidence of the second highest ranking knowledge snippet).  If this condition is met, implying that the top-ranking snippet is very likely correct, then GPT2-XL-RR is used for response generation. Otherwise, the ensemble method falls back to GPT2-XL, which can extract information from all top-k snippets, rather than only utilizing the single top snippet.

\begin{table}[!t]
  \centering
  \begin{threeparttable}
  \fontsize{10}{9}
  \selectfont
    \begin{tabular}{lccc}
        \toprule
         \textbf{Model} & \textbf{MRR@5} & \textbf{R@1} & \textbf{R@5}\cr
        \midrule
        Reproduced Baseline &0.830 &0.731 & 0.957\cr
        KnowleDgEFactor &\textbf{0.973} &\textbf{0.964} & \textbf{0.984}\cr
        \bottomrule
    \end{tabular}
  \end{threeparttable}
  \caption{Evaluation results of knowledge selection task on DSTC9 Track 1 Validation Set for all true knowledge-seeking turns. Line-1 is the reproduced GPT2-Baseline and line-2 is the performance of KnowleDgEFactor.}
  \label{tab:task2val}
\end{table}

\begin{table*}[!t]
  \centering
  \begin{threeparttable}
    \fontsize{10}{9}
    \selectfont
    \begin{tabular}{llccc}
    \toprule  
    \textbf{Model} &\textbf{Source of Task 1 Predictions}&\textbf{MRR@5} &\textbf{R@1} &\textbf{R@5} \cr 
    \midrule
    Official GPT2-Baseline & Official GPT2-Baseline&0.726 &0.620 & 0.877 \\
    KnowleDgEFactor & Task 1 Reproduced GPT2-Baseline& \textbf{0.853} & \textbf{0.827} & \textbf{0.896}\\
    \midrule
    KnowleDgEFactor & Ground Truth &0.903 &0.867 & 0.960\\
    \bottomrule  
    \end{tabular}
  \end{threeparttable}
  \caption{Evaluation results of knowledge selection task on the DSTC9 Track 1 Test Set. The 1st row is the released results on official GPT2-Baseline and the 2nd row shows KnowleDgEFactor's performance. The 3rd row shows the results operated on the ground-truth Task 1 prediction to evaluate our system independently.}
  \label{tab:task2test}
\end{table*}

\begin{table}[!t]
  \centering
  \fontsize{10}{9}
  \selectfont
    \begin{tabular}{lcccc}
        \toprule
        \multirow{2}{*}{\textbf{\#Candidates}}&
        \multicolumn{2}{c}{\textbf{Percentage}}\cr  
         \cmidrule(lr){2-3} \cmidrule(lr){4-5}
        &Validation Set&Test Set\cr
        \midrule
        1 & 84.1 & 74.0\\
        2 & 13.8 & 23.7\\
        3 & 1.5 & 2.2\\
        4 & 0 & 0.1\\
        5 & - & 0\\
        \bottomrule
    \end{tabular}
  \caption{Percentage of the true knowledge-seeking turns with different number of domain and entity candidates retrieved by Module 1.}
  \label{tab:task2module1}
\end{table}

\begin{table}[!t]
  \centering
  \begin{threeparttable}
  \fontsize{10}{9}
  \selectfont
    \begin{tabular}{lccc}
    \toprule
    \multirow{2}{*}{\textbf{True Domain}}&
    \multicolumn{3}{c}{\textbf{\#Correct (\%Correct)}}\cr  
     \cmidrule(lr){2-4} 
    &Domain&Entity&Knowledge\cr
    \midrule
    Hotel & 567 (98.8)& 545 (96.1)& 513 (94.1)\\
    Restaurant &599 (98.0)& 577 (96.3) & 554 (96.0) \\
    Taxi &183 (98.9)  & - & 138 (75.4)\\
    Train &346 (99.7) & - & 283 (81.8)\\
    Attraction &256 (97.0) & 243 (94.9) & 230 (94.7)\\
    \bottomrule  
    \end{tabular}
  \end{threeparttable}
  \caption{Performance of domain, entity and knowledge selection for top-1 knowledge snippet by KnowleDgEFactor on true knowledge-seeking turns of DSTC9 Track 1 Test Set. A turn is considered for entity accuracy calculation only if the predicted domain is correct and for knowledge accuracy calculation only if both domain and entity are correct.}
  \label{tab:task2analysis}
\end{table}

\section{Experiments}
\subsection{Task 1 -- Knowledge-seeking Turn Detection}
\subsubsection{Evaluation Metrics.} 
We use precision, recall and F-Measure as the metrics to evaluate the knowledge-seeking turn detection task \cite{gunasekara2020overview}.
\subsubsection{Experimental Settings.}
We use HuggingFace's implementation \cite{Wolf2019HuggingFacesTS} of BART (large) model for the Domain and Knowledge Classifier. The models were trained independently with a batch size of 120 and 3704 warmup steps. The models were trained for 4 epochs and the epoch with best performance on validation set was chosen. In our submitted system, we use $N_{dialog}=2$ with premise and hypothesis templates, i.e., both the system and user response for the Domain Classifier, and $N_{dialog}=1$ without any templates, i.e., only user response for the Knowledge Classifier. In a later improved knowledge classifier, we use $N_{dialog}=2$ with premise template. To test the generalizability on OOD user queries, we train and test the baseline and KDEAK on 4 versions of 2 disjoint sets of domains, with 2 domains in each, respectively.

\subsection{Task 2 -- Knowledge Selection}
\subsubsection{Evaluation Metrics.}
The performance of KnowleDgEFactor is measured in terms of standard information retrieval evaluation metrics, including recall and mean reciprocal rank \cite{kim2020domain}.

\subsubsection{Experiment Settings.}
 The PyTorch implementation of the BERT base model (uncased) in HuggingFace Transformers \cite{Wolf2019HuggingFacesTS} is utilized. All three models (BERT-DE, BERT-D, BERT-K) are fine-tuned independently with 10 epochs. The number of negative candidates is set as 4, 3 and 8 for BERT-DE, BERT-D and BERT-K. The maximum token lengths of dialogue and
knowledge are 256 and 128 for BERT-DE and BERT-D; 128 and 128 for BERT-K.

\subsection{Task 3 -- Knowledge-grounded Response Generation}
\subsubsection{Evaluation Metrics.}
Standard objective evaluation metrics are used for the system-generated response in comparison with the ground truth -- BLEU \cite{papineni2002bleu}, METEOR \cite{lavie2007meteor} and ROUGE \cite{lin2004automatic} \cite{lin2003automatic}.

\subsubsection{Experiment Settings.}

We fine-tuned the pre-trained GPT2 \textit{XL} on DSTC9 training set, and the loss function is the standard language modeling objective: cross-entropy between the generated response and the ground truth response. 
We set the input length limit as 128 tokens (i.e., words) for the dialog history and 256 tokens for the knowledge snippets. This means that we can typically fine-tune with 9 dialog turns and 4 snippets. The model is trained for 3 epochs with a size of 4. The gradient accumulation and gradient clipping with a max norm of $1.0$  were performed at every step. The optimizer was Adam and the learning rate was $e^{-6}$.

To achieve better performance with the generation model, we also compare different numbers of snippets to find the best setting that can provide enough information without introducing too much noise. Table \ref{tab:multi-snippet} presents the result on validation set with 1 to 5 snippets.
GPT2-XL and GPT2-XL-RR are evaluated on the Test Set in isolation and in combination in the ensemble system in Table \ref{tab:task3-test}.
\section{Results and Analysis}
\subsection{Task 1 -- Knowledge-seeking Turn Detection}
Table \ref{tab:KDEAK-performance} summarizes the results of Task 1. KDEAK outperforms the baseline on 3 out of 4 versions of the OOD F1-Score evaluations. The Domain Classifier shows 98.7\% accuracy on the DSTC9 Track 1 Val Set. Exploiting the pre-trained knowledge and rich hypothesis of the NLI model, KDEAK is more robust against unseen domains compared to non-NLI based GPT-2 baseline. It offers a transparent decision-making process at the domain, entity and information levels through its modular design. After the challenge, we improved the Knowledge Classifier which outperforms the baseline on DSTC9 Track 1 test set, by using both user and system response ($N_{dialog}=2)$ as the premise.
\subsection{Task 2 -- Knowledge Selection}
\subsubsection{Strength of a Factorized Approach.} 
The performance improvements on both DSTC9 Track 1 Validation (Table \ref{tab:task2val}) and Test (Table \ref{tab:task2test}) Sets over the baseline model demonstrate the advantage of a factorized approach to knowledge selection. One possible advantage of dividing the computation could be that the domain, entity and knowledge information from the dialogs is disentangled, and consequently the models of the three modules can respectively capture the traits about the three sub-tasks more easily. 84\% and 74\% of the true knowledge-seeking turns have only one entity or one domain $d_i \in D_w$ retrieved by Module 1 from each dialog of the Validation and Test Sets respectively (Table \ref{tab:task2module1}), and over 93\% of the top-1 retrieved entity is correct on both data sets, demonstrating the robustness and precision brought by the SMA and the BERT-DE.

\begin{figure}
\centering
\includegraphics[width=\columnwidth]{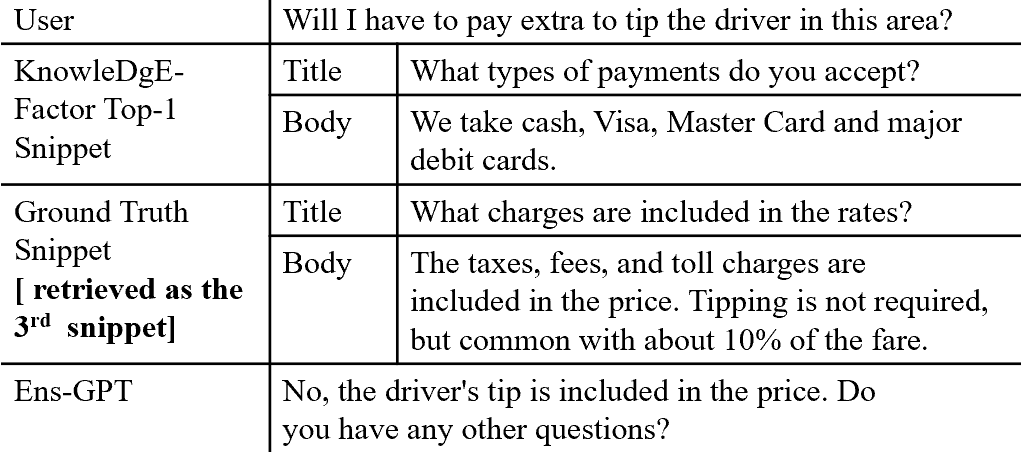}
\caption{Case study for error analysis}
\label{fig:task3case}
\end{figure}

\begin{table*}[hbtp]
\centering
\fontsize{10}{9}
\selectfont
\begin{tabular}{lcccccccc}
\toprule
\textbf{Model} & \textbf{BLEU-1} & \textbf{BLEU-2} & \textbf{BLEU-3} & \textbf{BLEU-4} & \textbf{METEOR} & \textbf{ROUGE-1} & \textbf{ROUGE-2} & \textbf{ROUGE-L} \\
 \midrule
  Baseline & 0.3031 & 0.1732 & 0.1005 & 0.0655 & 0.2983 & 0.3386 & 0.1364 & 0.3039\\
  
  GPT2-XL & 0.3550& 0.2297 & 0.1536 & \textbf{0.1048} & 0.3593 & 0.3972 & 0.1904 & \textbf{0.3571} \\
  
 GPT2-XL-RR & 0.3521 & \textbf{0.2336} & \textbf{0.1542} & 0.1042 & \textbf{0.3780} & 0.3957 & \textbf{0.1943} &0.3507 \\

 Ens-GPT & \textbf{0.3550} & 0.2300 & 0.1532 & 0.1040 & 0.3594 & \textbf{0.3976} & 0.1907 & 0.3570 \\
\bottomrule
\end{tabular}
\caption{Evaluation results of the ensemble system and its components on the DSTC9 Track 1 Test Set.}
\label{tab:task3-test}
\end{table*}

\begin{table}[!t]
  \centering
  \fontsize{10}{9}
  \selectfont
    \begin{tabular}{lcccc}
\toprule
 &\textbf{BLEU-1} & \textbf{BLEU-4} & \textbf{METEOR} & \textbf{ROUGE-L} \\
 \midrule
 T  & 0.3550 & 0.1040 & 0.3594 & 0.3570 \\
  \midrule
 1 & \textbf{0.3909} & \textbf{0.1192} & \textbf{0.3974} & \textbf{0.3922} \\
 2 & 0.3066 & 0.0664 & 0.2911 & 0.3157 \\
 3 & 0.2526 & 0.0257 & 0.2527 & 0.2586 \\
\bottomrule
\end{tabular}
\caption{Error analysis of the Ens-GPT's performance with rows T = Test set, 1 = Case 1, 2 = Case 2 and 3 = Case 3.}
\label{tab:task3-error-analysis}

\end{table}

\begin{table}[t]
\centering
\fontsize{10}{9}
\selectfont
\begin{tabular}{lcccc}
\toprule
\textbf{N}  &\textbf{BLEU-1}& \textbf{BLEU-4} & \textbf{METEOR}& \textbf{ROUGE-L} \\
 \midrule
1 &0.4184 & \textbf{0.1341} & 0.4220 & 0.4126 \\
2 & 0.4111 & 0.1194 & 0.4148 &   0.4034 \\
3 & 0.4133 & 0.1240 & 0.4173  & 0.4065 \\
4 & \textbf{0.4212} & 0.1292 &  \textbf{0.4270} & \textbf{0.4134}\\
5 & 0.4156 & 0.1266 & 0.4200 & 0.4075 \\
\bottomrule
\end{tabular}
\caption{Evaluation results with varying number of knowledge snippets $n \in \{1..5\}$  on the DSTC9 Track 1 Val Set. }
\label{tab:multi-snippet}
\end{table}

\subsubsection{Error Analysis.}
Despite the improved performance over the baseline, there is a noticeable decline of accuracy on the DSTC9 Track 1 Test Set as compared to that on the Validation Set. The drop can be attributed to the inability of Module 3 to recover the correspondence between the current user utterance and the knowledge snippets that are unseen during training. Table \ref{tab:task2analysis} records the domain-, entity- and knowledge-level accuracies of the top-1 selected knowledge snippet for all true knowledge-seeking turns. Although most of the selections are over 94\% accurate, it is shown that the errors mainly originate from the incorrect knowledge selection on the \textit{train} and \textit{taxi} domains, where we find that KnowleDgEFactor sometimes fails to distinguish between similar knowledge snippets. For example, 19 similar errors are found when the user asks about payment under the domain \textit{taxi}. Figure \ref{fig:task3case} shows an example of such erroneous instances where the correct knowledge snippet is ranked third. In this example, KnowleDgEFactor associates `pay' in the user query and `payments' in the title of the selected knowledge snippet without attending to the signaling word `tip'.

\subsection{Task 3 -- Knowledge-grounded Response Generation}

Empirical results in Table \ref{tab:multi-snippet} indicate that the use of an appropriate number of additional knowledge snippets (i.e., $n = 4$ in total) tends to result in improved performance compared to exclusive use of the top-ranking snippet. However, when $n\neq4$ performance degrades.

Table \ref{tab:task3-test} shows that all the methods, namely GPT2-XL and GPT2-XL-RR, as well as their ensemble, outperform the baseline (GPT2-small with a single knowledge snippet).  Also, the ensemble system outperforms GPT2-XL in 5/8 metrics.  However, comparison also shows that the ensemble method only outperforms GPT2-XL-RR in 3/8 evaluation metrics. This is the reverse of what was observed in the Validation Set and invites further investigation in future work.

\subsection{Integrated System Analysis.}
Misprediction in Task 1 directly affects Task 2's scores. Among the 1,981 true knowledge-seeking turns on the DSTC9 Track 1 Test data, 1,767 are correctly predicted by Task 1 and then processed by Task 2. There are 129 turns where Task 2 selects the wrong knowledge snippet, which together with the 214 errors in Task 1, account for recall@1 loss.

We also analyzed how the output of Task 2 output may influence performance in Task 3.  We categorize the output into 3 cases:  \textbf{Case 1}: ground truth knowledge snippet retrieved as the top snippet; \textbf{Case 2}: ground truth knowledge snippet appearing in the top-4 snippet list but not as the top-ranking snippet; \textbf{Case 3}: ground truth knowledge snippet not retrieved.
Table \ref{tab:task3-error-analysis} presents the performance of Ens-GPT on the Test Set under these three cases, which indicates that the quality of output from Task 2 greatly influences performance in Task 3. Also, adopting multi-knowledge snippets as input shows great importance as the model performs much better under  \textbf{Case 2} than \textbf{Case 3}.
As the example in Figure \ref{fig:task3case} shows, if the ground truth is retrieved as the third snippet, then the system Ens-GPT can still prevent the error propagation from Task 2 to 3 and answered the question correctly. 

We note that the approaches to Tasks 1 and 2 evolved to become convergent with some overlapping goals, but they are still different in certain fundamental aspects. In the future, we would like to develop a more streamlined approach, possibly combining Tasks 1 and 2 into a single sub-system.


\section{Conclusion}
 
We presented a pipeline of KDEAK, KnowleDgEFactor, and Ens-GPT, which achieves task-oriented dialog modeling with unstructured knowledge access, that can respond to users' request for information lying outside the database but in an external knowledge repository of FAQ-like snippets.

Task 1 (knowledge-seeking turn detection) is accomplished by a subsystem named KDEAK.  It formulates the problem as natural language inference and fully utilizes three information sources -- dialog history, database, and external knowledge. Domain and entity information determine the candidate pool of information snippets which are ranked based upon relevance to the user's query. Final classification is based on the source of most relevant information snippet.

Task 2 (knowledge selection) resorts to a 3-module KnowleDgEFactor subsystem formulated as a knowledge/document retrieval problem.  It is factorized into the sub-problems of domain and entity selection, domain probability estimation and knowledge probability estimation, which are handled by three modules. Knowledge snippets are ranked using the probabilities computed by the estimates of the modules.

Finally, Task 3 (knowledge-grounded response generation) is performed by Ens-GPT, in which multiple retrieved knowledge snippets are integrated to enrich knowledge and improve robustness of the generated response. The domain of the user query and the confidence of the retrieved snippets are used to determine which way to generate the response. 

Automatic evaluation metrics show that the final responses generated from integration of the three subsystems outperform the baseline significantly. 

Possible future directions may include extension towards open-domain knowledge-grounded conversations \cite{Gopalakrishnan2019}, enhancing robustness towards recognition errors for speech inputs \cite{gopalakrishnan2020neural} and creating an engaging user experience.

\section{Acknowledgments}
This work is partially supported by the Centre for Perceptual and Interactive Intelligence, a CUHK InnoCentre.  We thank Dr. Pengfei Liu, a graduate of the Department of Systems Engineering \& Engineering at CUHK, for constructive comments and suggestions.

\bibliography{DSTC9bib}

\end{document}